\documentclass[final,3p,times,twocolumn]{elsarticle}

\usepackage{amssymb}
\usepackage{hyperref}
\usepackage{graphicx}
\usepackage{subcaption}
\usepackage{lscape}
\usepackage{amsmath}
\usepackage{multirow}
\usepackage[figuresright]{rotating}
\usepackage{algorithm} 
\usepackage{algpseudocode} 
\usepackage{comment}

\usepackage{natbib}
\setcitestyle{authoryear,open={(},close={)},citesep={;}} 

\usepackage{calc}
\begin{document}
	
\begin{frontmatter}
		
	\title{Active Laser-Camera Scanning for High-Precision Fruit Localization in Robotic Harvesting: System Design and Calibration}
		
	\author[label1]{Kaixiang Zhang}
	\author[label2]{Pengyu Chu}
	\author[label2]{Kyle Lammers}
	\author[label1]{Zhaojian Li}
	\author[label3]{Renfu Lu}
		
	\address{*Zhaojian Li (lizhaoj1@egr.msu.edu) is the corresponding author}
	\address[label1]{Department of Mechanical Engineering, Michigan State University, East Lansing, MI 48824, USA}
	\address[label2]{Department of Electrical and Computer Engineering, Michigan State University, East Lansing, MI 48824, USA}
	\address[label3]{United States Department of Agriculture, Agricultural Research Service, East Lansing, MI 48824, USA}
		
\begin{abstract}
	Robust and effective fruit detection and localization is essential for robotic harvesting systems. While extensive research efforts have been devoted to improving fruit detection, less emphasis has been placed on the fruit localization aspect, which is a crucial yet challenging task due to limited depth accuracy from existing sensor measurements in the natural orchard environment with variable lighting conditions and foliage/branch occlusions. In this paper, we present the system design and calibration of an Active LAser-Camera Scanner (ALACS), a novel perception module for robust and high-precision fruit localization. The hardware of ALACS mainly consists of a red line laser, an RGB camera, and a linear motion slide, which are seamlessly integrated into an active scanning scheme where a dynamic-targeting laser-triangulation principle is employed. A high-fidelity extrinsic model is developed to pair the laser illumination and the RGB camera, enabling precise depth computation when the target is captured by both sensors. A random sample consensus-based robust calibration scheme is then designed to calibrate the model parameters based on collected data. Comprehensive evaluations are conducted to validate the system model and calibration scheme. The results show that the proposed calibration method can detect and remove data outliers to achieve robust parameter computation, and the calibrated ALACS system is able to achieve high-precision localization with millimeter-level accuracy. 
\end{abstract}
		
		
\end{frontmatter}

\section{Introduction}
\label{sec:intro}
	
With the growing global population, the agriculture industry has been pushing to adopt mechanization and automation for increasing, sustainable food production at lower economic and environmental costs. While such technologies have been deployed for field crops such as corn and wheat, the fruit sector (e.g., apple, citrus and pear) still heavily relies on seasonal, manual labor. In many advanced economies, the availability of labor for farming has been on steady decline, while the cost of labor has increased significantly. Moreover, tasks like manual harvesting involve extensive body motion repetitions and awkward postures (especially when picking fruits at high places or deep in the canopy, and repeatedly ascending and descending on ladders with heavy loads), which put workers at risk for ergonomic injuries and musculoskeletal pain~\citep{Fathallah2010AE}. Considering the aforementioned issues, robotic harvesting is thus considered to be a promising solution for sustainable fruit production and has received increasing attention in recent years.
	
Research on robotic harvesting technology has been ongoing for several decades, and different robotic systems have been attempted for semi-automated or fully automated fruit harvesting~\citep{De2011,Mehta2014CEA,DeASABE2015,SilwalJFR2017,Xiong2018CEA,Williams2019BE,HohimerASABE2019,ZhangASABE2020,Zhang2021MECH,BuCEA2022,Zhang2022IROS}. A typical robotic harvesting system consists of a perception module, a manipulator, and an end-effector. Specifically, the perception module exploits onboard sensors (e.g., cameras and LiDARs) to detect and localize the fruit. Once the fruit position is determined by the perception system, the manipulator is controlled to reach the target fruit, and then a specialized end-effector (e.g., gripper or vacuum tube) is actuated to detach the fruit. Therefore, the development of a robotic harvesting system requires multi-disciplinary advancements to enable a variety of synergistic functionalities. Among the various tasks, fruit detection and localization is the first and foremost one to support robotic manipulation and fruit detachment. Specifically, the fruit detection function aims at segmenting fruits from the complex background, while the localization is to calculate the spatial positions of the detected fruits. Due to variable lighting conditions, color variations of fruits with different degrees of ripeness and varietal differences, and fruit occlusions by foliage and branches, developing sensing modules and perception algorithms capable of robust and effective fruit detection and localization in the real orchard environment poses significant technical challenges.
	
To date, extensive studies have been devoted to efficient and robust fruit detection, which is most commonly accomplished using color images captured by RGB cameras. In general, these approaches can be classified into two categories: feature-based and deep learning-based. The feature-based methods~\citep{Bulanon2002BE,ZhaoIROS2005,Wachs2010PA,Zhou2012PA,Nguyen2016BE,Lin2020PA} use differences among predefined features (e.g., color, texture, and geometric shape) to identify the fruit, and various conventional computer vision techniques (e.g., Hough transform-based circle detection method, optical flow method, and Ostu adaptive threshold segmentation) are used for feature extraction. Such methods perform well under certain simple harvesting scenarios but are susceptible to varying lighting conditions and heavy occlusions. This is because the extracted features are defined artificially and they are not universally adaptable and may lack generalization capabilities in distinguishing target fruits when the harvesting scene changes~\citep{Li2022RS}. Different from feature-based methods, deep learning-based methods exploit convolutional neural networks to extract abstract features from color images, making them suitable for complex recognition problems. Deep learning-based object recognition algorithms have seen tremendous success in recent years, and a variety of network structures, i.e., region convolution neural network (RCNN)~\citep{Girshick2014CVPR}, Faster RCNN~\citep{Ren2017PAMI}, Mask RCNN~\citep{He2020PAMI,ChuPRL2021}, You Only Look Once (YOLO)~\citep{Redmon2018arXiv,TianCEA2019,KangCEA2020}, and Single Shot Detection (SSD)~\citep{Liu2016ECCV}, have been studied and extended for fruit detection. Specifically, RCNN based approaches employ a two-stage network architecture, in which a region proposal network (RPN) is used to search the region of interest and a classification network is used to conduct bounding box regression. As opposed to two-stage networks, YOLO and SSD based one-stage networks merge the RPN and classification branch into a single convolution network architecture, which enjoys improved computation efficiency.
	
Once the fruits are recognized and a picking sequence is determined (see e.g., \cite{Zhang2022IROS}), 3-dimensional (3D) localization needs to be conducted to compute the spatial coordinates of a target fruit. Accurate fruit localization is crucial since erroneous localization will cause the manipulator to miss the target and subsequently degrade the harvesting performance of the robotic system. Various sensor configurations and techniques have been used for fruit localization~\citep{Gongal2015,Gene2019BE,FuCEA2020,Neupane2021AGR,Kang2022CEA}. One example is (passive) stereo vision systems, which exploit two-camera layout and triangulation optical measurement principle to obtain depth information. For such systems, the relative geometric pose of the two cameras needs to be carefully designed and calibrated, and sophisticated algorithms are required to search common features in two dense RGB images for stereo matching. Therefore, the main disadvantages of stereo vision systems are that the generation of depth information is computationally expensive and the performance of stereo matching is inevitably affected by occluded pixels or varying lighting conditions that are common in the natural orchard environment.  
	
Consumer RGB-D cameras are another type of depth measurement sensors that have recently been employed to localize fruits~\citep{Xiong2019CEA,Tian2019ACCESS,Arad2020JFR,Kang2020ACCESS}. Different from passive stereo vision systems that purely rely on natural light, the RGB-D sensors include a separate artificial illumination source to aid the depth computation. According to the methods on how the depth measurements are computed, RGB-D cameras can be divided into three categories: structured light (SL), time of flight (ToF), and active infrared stereo (AIRS)~\citep{FuCEA2020}. An SL-based RGB-D sensor usually consists of a light source and a camera system. The light source projects a series of light patterns onto the workspace, and the depth information can then be extracted from the images based on the deformation of the light pattern. The first-generation Kinect (Microsoft Corp., Redmond, WA, USA) and the RealSense F200 and SR300 (Intel Corp., Santa Clara, CA, USA) are representative consumer sensors that operate with SL, and they have been utilized in different agricultural applications~\citep{Lehnert2017RAL,Liu2018SEN,Milella2019CEA}. The ToF-based RGB-D sensors use an infrared light emitter to emit light pulses onto the scene. The distance between the sensor and the object is calculated based on the known speed of light and the round trip time of the light signal. One important feature of the ToF systems is that their depth measurement precision does not deteriorate with distance, which makes them suitable for harvesting applications requiring a long perception range. Moreover, the AIRS-based RGB-D sensors are an extension of the conventional passive stereo vision system. They combine an infrared stereo camera pair with an active infrared light source to improve the depth measurement under low-texture environment. One of the most widely used AIRS sensors in fruit localization is the RealSense D400 family (Intel Corp., Santa Clara, CA, USA). 
Despite some successes, the sensors mentioned above may have limited and unstable performance in the natural orchard environment.
For example, the SL-based sensors are sensitive to the natural light condition and to the interference of multiple patterned light sources. The ToF systems are vulnerable to scattered light and multi-path interference, and usually provide lower resolution of depth images compared to other RGB-D cameras. Similar to passive stereo vision systems, the AIRS-based sensors encounter stereo matching issues, which can lead to flying pixels or over-smoothing around the contour edges~\citep{FuCEA2020}. In addition, the performance of these sensors could deteriorate significantly when target fruits are occluded by leaves and branches, due to low or limited density of the illuminating light patterns or point cloud.
	
It is thus clear that both the stereo vision systems and the RGB-D sensors have inherent depth measurement limitations in providing precise fruit localization information that is necessary for effective robotic harvesting systems. Towards this end, we devise a novel perception module, called Active LAser-Camera Scanner (ALACS), to improve fruit localization accuracy and robustness for ready deployment in  apple harvesting robots. 
In this paper, we present the system design and calibration scheme of ALACS, and the main contributions of this paper are highlighted as follows.
\begin{enumerate}
	\item A hardware system consisting of a red line laser, an RGB camera, and a linear motion slide, coupled with an active scanning scheme, is developed for fruit localization based on the laser-triangulation principle.
	\item A high-fidelity extrinsic model is developed to capture 3D measurements by matching the laser illumination source with the RGB pixels. A robust calibration scheme is then developed to calibrate the model parameters by leveraging random sample consensus (RANSAC) techniques to detect and remove data outliers.
	\item The effectiveness of the developed model and calibration scheme is evaluated through comprehensive experiments.
\end{enumerate}

This is the first effort that, to the best of our knowledge, combines a line laser with a camera to accomplish millimeter-level localization performance. While ALACS is primarily developed and tested for the apple harvesting application, it can be readily extended and adopted for other tree fruits.
	
The rest of the paper is organized as follows. Section \ref{sec:overview} provides an overview of our newly-developed robotic apple harvesting system. Section \ref{sec:system} presents the system design of the ALACS. The extrinsic model for 3D measurement characterization and the corresponding robust calibration scheme are introduced in Section \ref{sec:calibration}. Simulation and experimental results are presented in Section~\ref{sec:results}. Finally, conclusions are drawn in Section ~\ref{sec:conclu}.
	
\section{Overview of the Robotic Apple Harvesting System} \label{sec:overview}
	
	\begin{figure*}[!h]
		\centering
		\begin{subfigure}[b]{0.455\textwidth}
			\centering
			\includegraphics[width=1\textwidth]{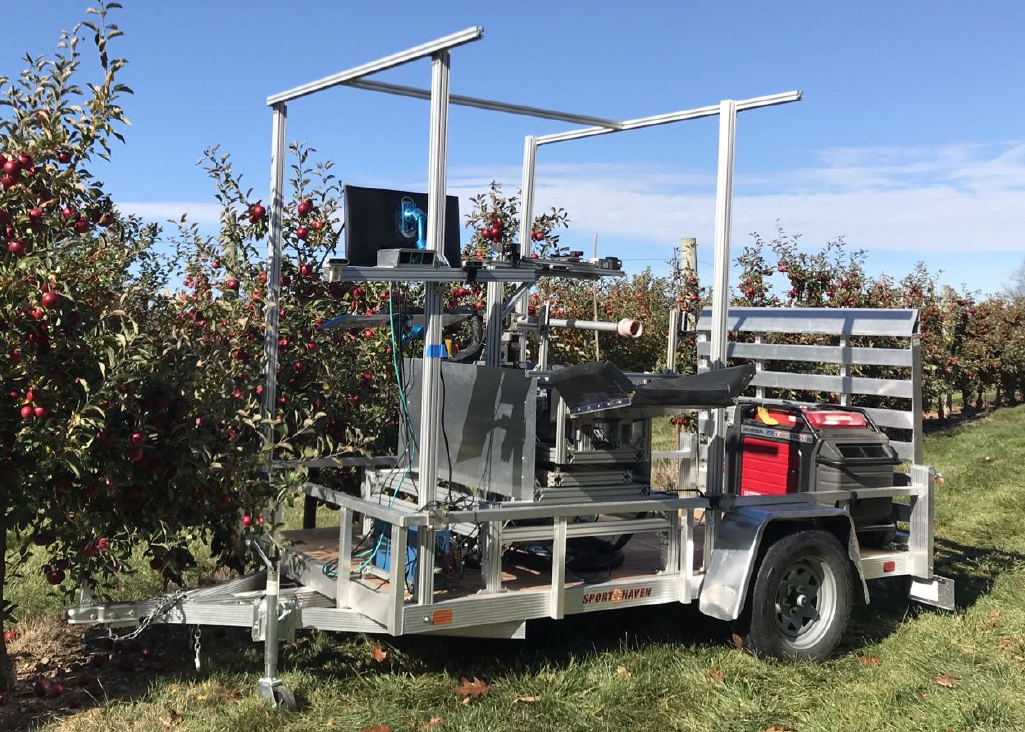}
			\caption{}
			\label{fig_robot}
		\end{subfigure}
		\hfill
		\begin{subfigure}[b]{0.495\textwidth}
			\centering
			\includegraphics[width=1\textwidth]{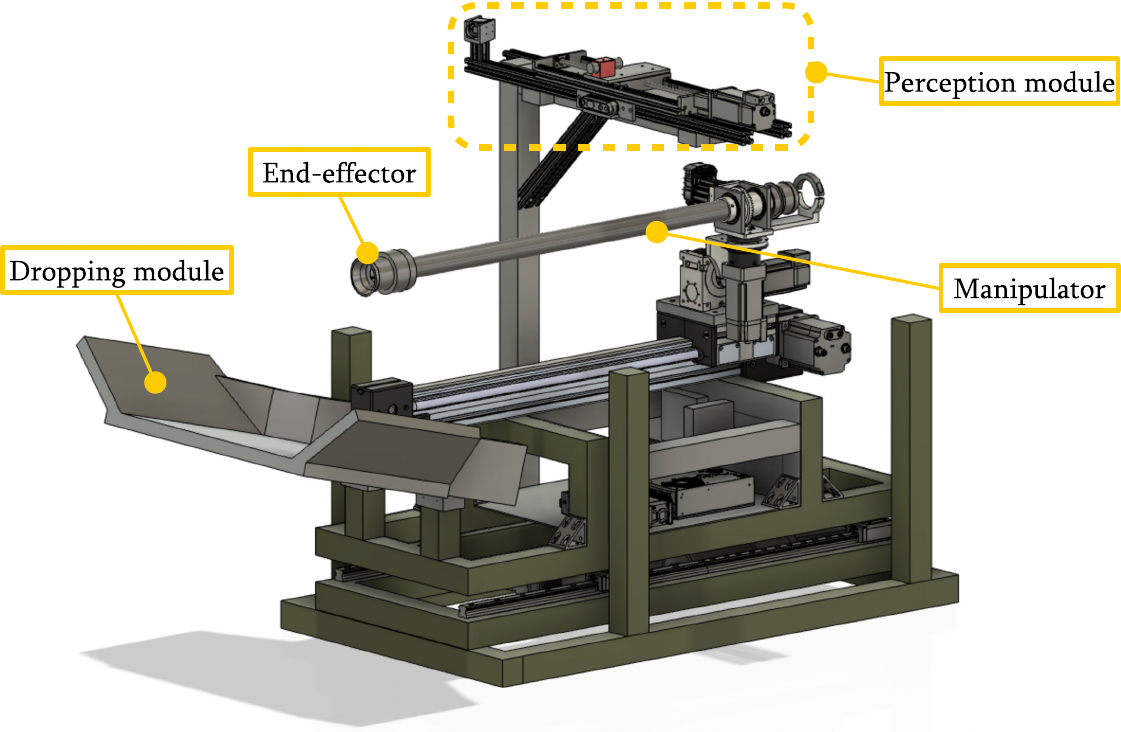}
			\caption{}
			\label{fig_robotCAD}
		\end{subfigure}
		\caption{The developed robotic apple harvesting system. (a) Image of the whole system operating in the orchard environment. (b) Main components of the robotic system.}
		\label{fig_system}
	\end{figure*}
	
In this section, we first briefly introduce our robotic apple harvesting platform, into which the ALACS is integrated. As shown in Figure~\ref{fig_system}, the robotic platform consists of four main components: a perception module, a 4 degree-of-freedom manipulator, a soft vacuum-based end-effector, and a dropping module. The robotic system is mounted on a trailer base to facilitate movement in the orchard environment. An industrial computer is utilized to coordinate the perception module, the manipulator, and all communication devices. The entire software is fully integrated using the robot operating system (ROS), where different software components are primarily communicated via custom messages.
	
The following introduces the steps that our system takes to harvest an apple. At the beginning of each harvesting cycle, the perception module is activated to detect and localize the fruits within the manipulator's workspace. 
Given the 3D apple location, the planning algorithm is used to generate a reference trajectory, and the control module then actuates the manipulator to follow this reference trajectory to approach the fruit. After successfully attaching the fruit to the end-effector, a rotation mechanism is triggered to rotate the end-effector by a certain angle, and then the manipulator is driven to pull and detach the apple. Finally, the manipulator retracts to a dropping spot and releases the fruit.
According to the aforementioned picking procedure, it can be seen that the fruit detection and localization is a key task in automated apple harvesting. 
Our previous system prototypes~\citep{Zhang2021MECH,Zhang2022IROS} utilized RGB-D cameras to facilitate fruit detection and localization. However, laboratory and field tests found that the commercial RGB-D cameras could not provide accurate depth information of the target fruits under leaf/branch occlusions and/or challenging lighting conditions. Inaccurate apple localization has been identified as one of the primary causes for harvesting failure. To enhance the apple localization accuracy and robustness, we designed a new perception unit (called ALACS), which seamlessly integrates the line laser with RGB image for active sensing.
	
\section{Design of the Active Laser-Camera Scanner}
\label{sec:system}
	
\begin{figure}[!h]
		\centering
		\includegraphics[width=0.40\textwidth]{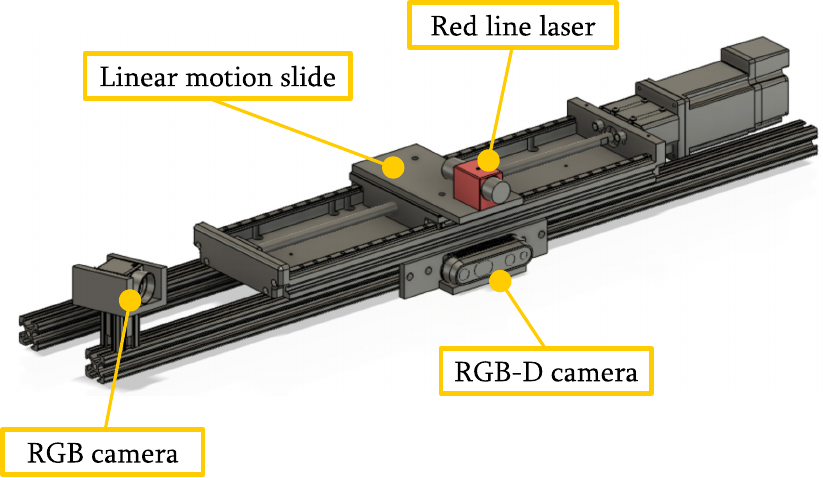}
		\caption{CAD model of the perception module.}
		\label{fig_perception}
	\end{figure}

As shown in Figure~\ref{fig_perception}, the perception module of the robotic apple harvesting system includes an Intel RealSense D435i RGB-D camera (Intel Corp., Santa Clara, CA, USA) and a custom ALACS unit. The RGB-D camera is mounted on a horizontal frame that is above the manipulator to provide a global view of the scene. The ALACS unit is comprised of a red line laser (Laserglow Technologies, North York, ON, Canada), a FLIR RGB camera (Teledyne FLIR, Wilsonville, OR, USA), and a linear motion slide. The line laser is mounted on top of the linear motion slide that enables the laser to move left and right horizontally with a full stroke of 20 cm. Meanwhile, the FLIR RGB camera is installed at the rear end of the linear motion slide with a relative angle to the laser. The hardware configuration of ALACS is designed to facilitate depth measurements using the principle of laser triangulation. The laser triangulation-based technique captures depth measurements by pairing a laser illumination source with a camera, which has been widely used in industry applications for precision 3D object profiling. It should be noted that the ALACS unit is different from the conventional laser triangulation sensors. For conventional laser triangulation sensors, the relative position between the laser and the camera is fixed (i.e., both of them are either stationary or moving simultaneously). For ALACS, the camera is fixed while the laser position can be adjusted with the linear motion slide. 
	
The RGB-D camera and the ALACS unit are fused synergistically to achieve apple detection and localization. Specifically, the fusion scheme includes two steps. In the first step, the images captured by the RGB-D camera are fed into a deep learning approach for fruit detection (see~\cite{Chu2023arXiv}), and the target apple location is then roughly calculated with the depth measurements provided by the RGB-D camera. In the second step, by using the rough apple location, the ALACS unit is triggered to actively scan the target apple, and an ameliorative apple position is obtained. As shown in Figure~\ref{fig_principle}, the basic working principle of ALACS is to project the laser line onto the target fruit and then use the image information and triangulation technique to localize the fruit.
The perception strategy of the ALACS unit is designed as follows:
	
\begin{figure}[!h]
	\centering
	\includegraphics[width=0.48\textwidth]{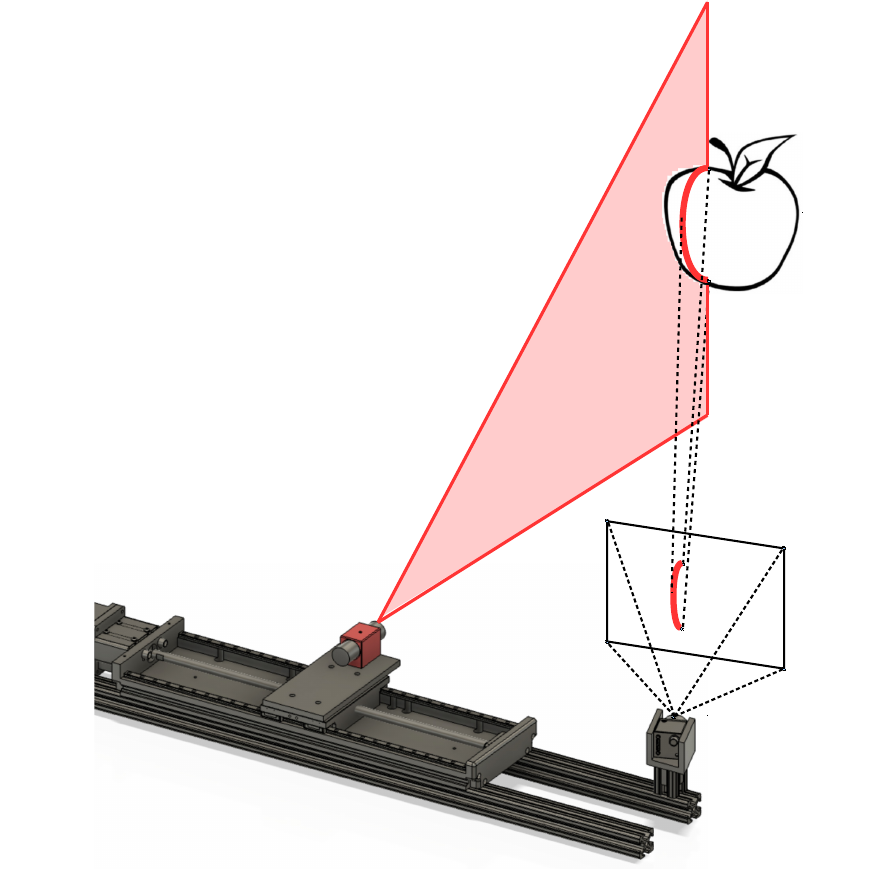}
	\caption{Fundamental working principle of the ALACS unit.}
	\label{fig_principle}
\end{figure}
	
\begin{enumerate}
	\item \textbf{Initialization.} The linear motion slide is actuated to regulate the laser towards an initial position, ensuring that the red laser line is projected on the left half region of the target apple. The initial laser position is obtained by transforming the rough target apple location provided by the RGB-D camera into the coordinate frame of the ALACS unit. 
		
	\item \textbf{Interval scanning.} When the laser reaches the initial position, the FLIR camera is activated to capture an image. The linear motion slide then travels to the right by four centimeters in one centimeter increments, pausing at each increment to allow the FLIR camera to take an image. A total of five images are acquired through this scanning procedure, with the laser line projected on various positions in each image. 
	The purpose of utilizing such scanning strategy is to mitigate the impact of occlusion, since the laser line provides high spatial-resolution localization information for the target fruit. More precisely, when the target apple is partially occluded by foliage, moving the laser to multiple positions can reduce the likelihood that the laser lines will be entirely blocked by the obstacle.
		
	\item \textbf{Refinement of 3D position.} For each image captured by the FLIR camera, the laser line projected on the target apple surface is extracted and then used to generate a 3D location candidate. Computer vision approaches and laser triangulation-based techniques are exploited to accomplish laser line extraction and position candidate computation, respectively. Five position candidates will be generated as a result, and a holistic evaluation function is used to select one of the candidates as the final target apple location. 
\end{enumerate}
	
To accomplish the aforementioned fruit localization scheme, laser line extraction and position candidate computation are two key tasks. The laser line extraction is achieved by leveraging computer vision techniques, and a detailed description on the extraction algorithm can be found in our recent work~\citep{ZhangJFR2023}. To facilitate the computation of fruit 3D positions, a high-fidelity model is derived based on the principle of laser triangulation, and a robust calibration scheme is designed. The following will detail the development of the high-fidelity model and calibration scheme.
	
\section{Extrinsic Model and Calibration} \label{sec:calibration}
	
\subsection{Modeling of the ALACS Unit}
\begin{figure}[!b]
	\centering
	\includegraphics[width=0.40\textwidth]{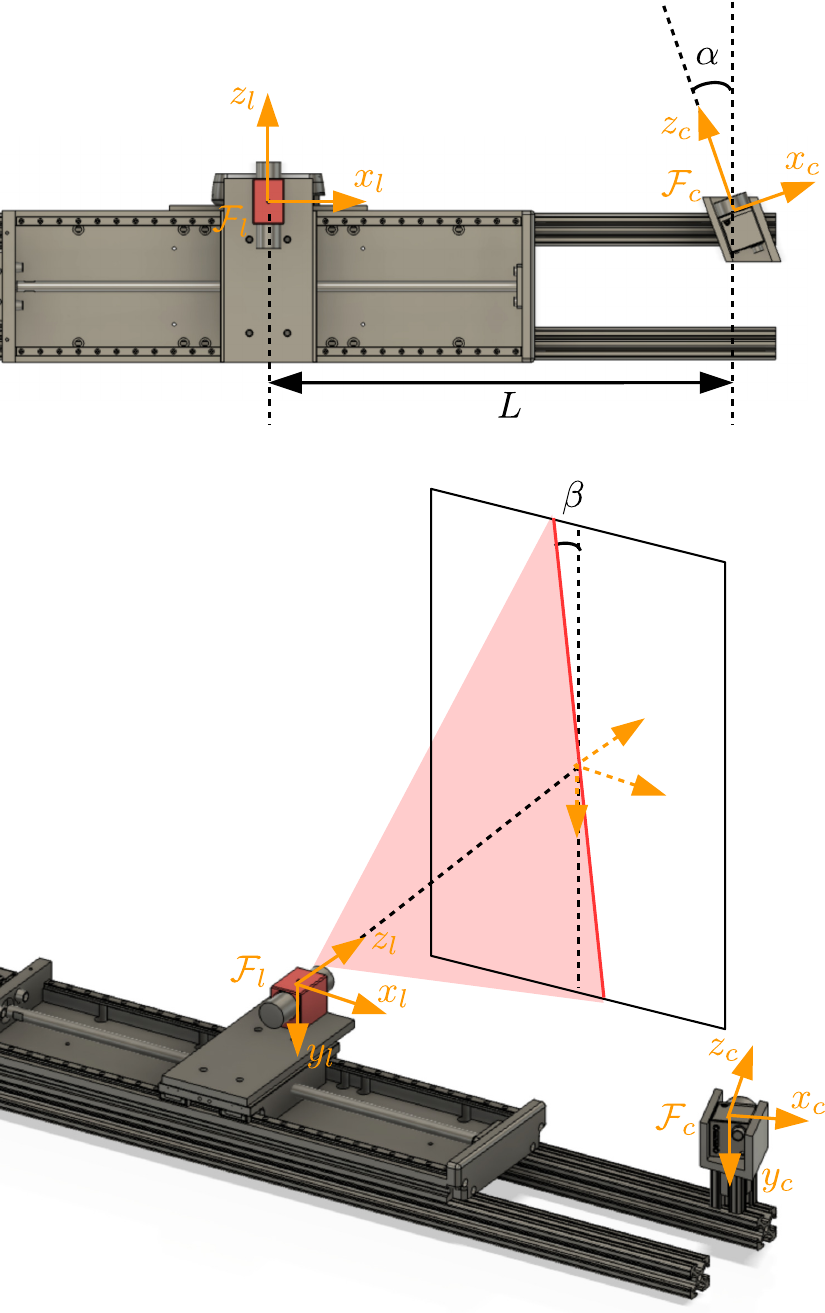}
	\caption{Coordinate frames and extrinsic parameters of the ALACS unit.}
	\label{fig_extrinsic}
\end{figure}
The basic idea of laser triangulation-based technique is to capture depth measurements by pairing a laser illumination source with a camera. Both the laser beam and the camera are aimed at the target object, and based on the extrinsic parameters between the laser source and the camera sensor, the depth information can be collected with trigonometry. As shown in Figure~\ref{fig_extrinsic}, $\mathcal{F}_{l}$ and $\mathcal{F}_{c}$ are denoted as the laser frame and camera frame, respectively. $\alpha \in \mathbb{R}$ is the rotating angle along the $y_{l}$-axis between $\mathcal{F}_{l}$ and $\mathcal{F}_{c}$. $L\in \mathbb{R}$ is the horizontal distance (i.e., the translation along the $x_{l}$-axis) between $\mathcal{F}_{l}$ and $\mathcal{F}_{c}$. $\beta\in \mathbb{R}$ is the angle between the laser plane and the $(y_{l}, z_{l})$ plane of $\mathcal{F}_{l}$. $\alpha$, $L$, and $\beta$ are considered as the extrinsic parameters between the laser illumination source and the camera, which are essential for deriving the high-fidelity model of the ALACS unit. In the following, we first introduce the pin-hole model of the camera and then present the model of ALACS.
	
Let $p_{i}$ be a point located at the intersection of the laser line and the object. The 3D position of $p_{i}$ under the camera frame $\mathcal{F}_{c}$ is denoted by $p_{c,i} = \begin{bmatrix}
	x_{c,i}, y_{c,i}, z_{c,i}
\end{bmatrix}^{\top}\in \mathbb{R}^{3}$. The corresponding normalized coordinate $\bar{p}_{c,i}\in \mathbb{R}^{3}$ is defined by
\begin{equation} \label{equ:bar_pc}
	\bar{p}_{c,i} = \begin{bmatrix}
		\bar{u}_{c,i}, \bar{v}_{c,i}, 1
	\end{bmatrix}^{\top} = \begin{bmatrix}
		\frac{x_{c,i}}{z_{c,i}}, \frac{y_{c,i}}{z_{c,i}}, 1
	\end{bmatrix}^{\top}.
\end{equation}  
Denote $m_{c,i}=\begin{bmatrix}
	u_{c,i}, v_{c,i}, 1
\end{bmatrix}^{\top}\in \mathbb{R}^{3}$ as the pixel coordinate of $p_{i}$ on the image plane. Then, the following pin-hole camera model can be used to describe the projection from $\bar{p}_{c,i}$ to $m_{c,i}$:
\begin{equation}
	m_{c,i} = \varpi (K\bar{p}_{c,i}),
\end{equation} 
where $\varpi (\cdot)$ is the camera distortion model and $K \in \mathbb{R}^{3\times 3}$ is the camera intrinsic matrix. Both $\varpi (\cdot)$ and $K$ can be obtained via standard calibration approaches, and thus once $m_{c,i}$ is detected from the image, the normalized coordinate $\bar{p}_{c,i}$ can be calculated by
\begin{equation} \label{equ:bar_pc_mc}
	\bar{p}_{c,i} = K^{-1} \varpi^{-1}(m_{c,i}).
\end{equation}
	
We now derive the high-fidelity model for the ALACS unit. Denote $p_{l,i} = \begin{bmatrix}
	x_{l,i}, y_{l,i}, z_{l,i}
\end{bmatrix}^{\top}\in \mathbb{R}^{3}$ as the 3D position of $p_{i}$ under the laser frame $\mathcal{F}_{l}$. According to the relative pose between $\mathcal{F}_{l}$ and $\mathcal{F}_{c}$ (see Figure~\ref{fig_extrinsic}), it can be concluded that
\begin{equation} \label{equ:pc_pl}
	\begin{bmatrix}
		x_{c,i} \\ y_{c,i} \\ z_{c,i}
	\end{bmatrix} = \begin{bmatrix}
		\cos(\alpha) & 0 & \sin(\alpha) \\
		0 & 1 & 0 \\
		-\sin(\alpha) & 0 & \cos(\alpha)
	\end{bmatrix} \begin{bmatrix}
		x_{l,i} \\ y_{l,i} \\ z_{l,i}
	\end{bmatrix} + \begin{bmatrix}
		-L\cos(\alpha) \\ 0 \\ L\sin(\alpha)
	\end{bmatrix}.
\end{equation}
In addition, as there is an angle, i.e., $\beta$, between the laser plane and the $(y_{l}, z_{l})$ plane of $\mathcal{F}_{l}$, we have 
\begin{equation} \label{equ:xl_yl}
	x_{l,i} = -y_{l,i} \tan(\beta).
\end{equation} 
Based on \eqref{equ:pc_pl} and \eqref{equ:xl_yl}, the following expression can be derived:
\begin{equation} \label{equ:tan}
	\tan(\alpha) = \frac{x_{c,i}+L\cos(\alpha)+y_{c,i}\cos(\alpha)\tan(\beta)}{z_{c,i}-L\sin(\alpha)-y_{c,i}\sin(\alpha)\tan(\beta)}.
\end{equation}
It can be concluded from \eqref{equ:bar_pc} that $x_{c,i} = z_{c,i}\bar{u}_{c,i}$ and $y_{c,i} = z_{c,i}\bar{v}_{c,i}$. After submitting these two relations into \eqref{equ:tan}, we can derive that
\begin{equation} \label{equ:zc}
	\begin{aligned}
		z_{c,i} = \frac{L}{\sin(\alpha) - \bar{u}_{c,i}\cos(\alpha) - \bar{v}_{c,i}\tan(\beta)}.
	\end{aligned}
\end{equation}
Using \eqref{equ:zc} and the facts that $x_{c,i} = z_{c,i}\bar{u}_{c,i}$ and $y_{c,i} = z_{c,i}\bar{v}_{c,i}$, we have
\begin{equation} \label{equ:xc_yc}
	\begin{aligned}
		x_{c,i} &= \frac{L\bar{u}_{c,i}}{\sin(\alpha) - \bar{u}_{c,i}\cos(\alpha) - \bar{v}_{c,i}\tan(\beta)}, \\
		y_{c,i} &= \frac{L\bar{v}_{c,i}}{\sin(\alpha) - \bar{u}_{c,i}\cos(\alpha) - \bar{v}_{c,i}\tan(\beta)}.
	\end{aligned}
\end{equation}
\eqref{equ:zc} and \eqref{equ:xc_yc} are the high-fidelity model that reveals the 3D measurement mechanism of the ALACS unit. Specifically, given the pixel coordinate $m_{c,i}$, $\bar{p}_{c,i}$, i.e., $\bar{u}_{c,i}$ and $\bar{v}_{c,i}$, can be computed via~\eqref{equ:bar_pc_mc}. Then, model~\eqref{equ:zc} and~\eqref{equ:xc_yc} can be exploited to calculate the 3D position $p_{c,i} = \begin{bmatrix}
	x_{c,i}, y_{c,i}, z_{c,i}
\end{bmatrix}^{\top}$ provided that the extrinsic parameters $\alpha$, $L$, and $\beta$ are well calibrated.

\subsection{Robust Calibration Scheme} \label{subsec:calibrationScheme}
The extrinsic parameters $\alpha$, $L$, and $\beta$ play a crucial role in facilitating the 3D measurement of the ALACS unit. In this subsection, we focus on introducing how we perform robust calibration on the extrinsic parameters $\alpha$, $L$, and $\beta$.  
Note that $\alpha$ and $\beta$ are constants, while $L$ is variable as the linear motion slide can move to different positions. During the calibration procedure, the linear motion slide is fixed at an initial position, and the corresponding horizontal distance between laser and camera is denoted by $L_{0}\in \mathbb{R}$. $\alpha$, $\beta$, and $L_{0}$ (i.e., the initial value of $L$) are obtained via offline calibration. Then, when the linear motion slide is moving, $L$ can be updated online based on its initial value $L_{0}$ and the movement distance of the linear motion slide.
	
The calibration procedure includes two steps. In the first step, multiple sets of data $s_{i} =\begin{bmatrix}
	\bar{u}_{c,i}, \bar{v}_{c,i}, z_{c,i}
\end{bmatrix}^{\top}\in \mathbb{R}^{3}$ ($i=1, 2, \cdots, n)$ are collected from recorded images. The second step then formulates an optimization problem by using the collected data and the model~\eqref{equ:zc} to compute the extrinsic parameters. The following details these two steps in sequence. 
	
\begin{figure*}[!h]
	\centering
	\begin{subfigure}[b]{0.34\textwidth}
		\centering
		\includegraphics[width=1\textwidth]{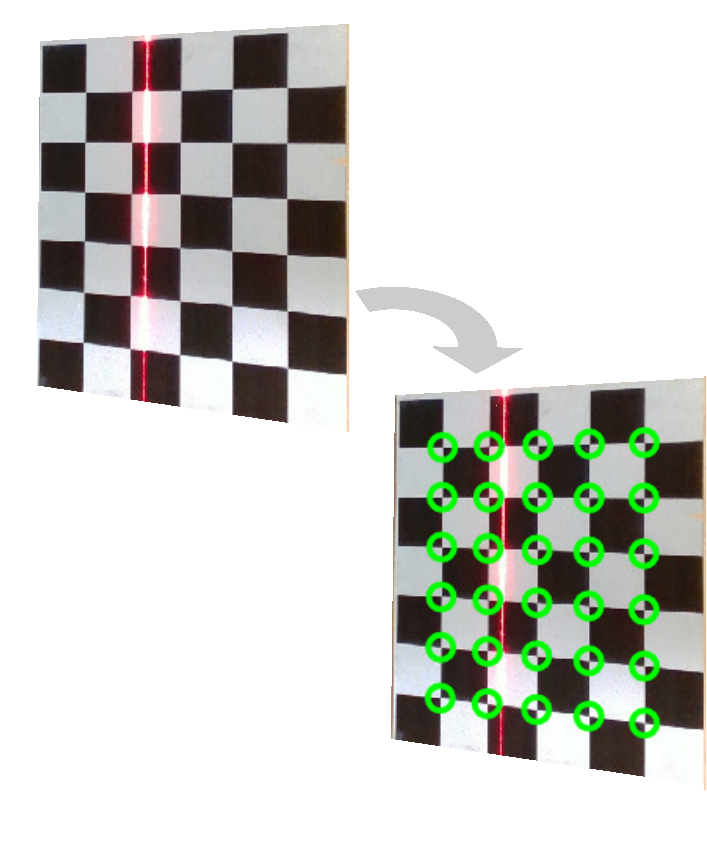}
		\caption{}
		\label{fig_corner}
	\end{subfigure}
	\hfill
	\begin{subfigure}[b]{0.2793\textwidth}
		\centering
		\includegraphics[width=1\textwidth]{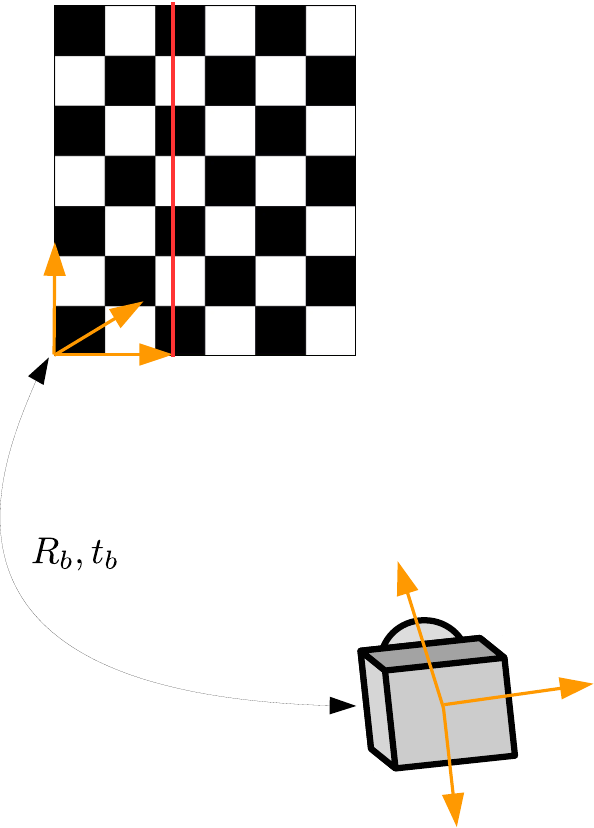}
		\caption{}
		\label{fig_reconstruction}
	\end{subfigure}
	\hfill
	\begin{subfigure}[b]{0.264\textwidth}
		\centering
		\includegraphics[width=1\textwidth]{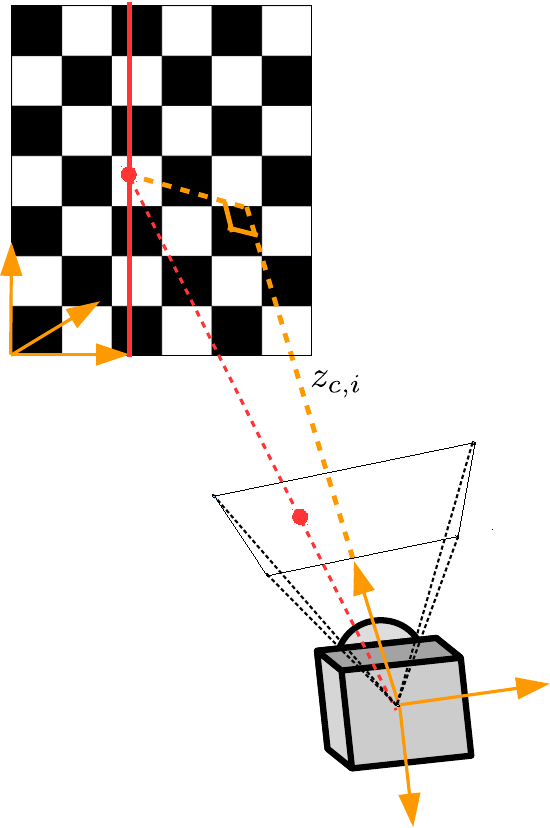}
		\caption{}
		\label{fig_depth}
	\end{subfigure}
	\caption{Scheme to compute $z_{c,i}$. (a) Corner detection. (b) Pose reconstruction. (c) Computation of $z_{c,i}$.}
	\label{fig_computation}
\end{figure*}

The hardware setup for image and data collection is shown in Figure~\ref{fig_setup}, where a planar checkerboard is placed in front of the ALACS unit so that the laser line will be projected on it. We use the planar checkerboard as the calibration pattern to facilitate the data collection. Specifically, given an image that covers the whole checkerboard, the pixel coordinates of laser points projected on the checkerboard are extracted based on the color feature. Once pixel coordinate $m_{c,i}$ is obtained, the corresponding normalized coordinate $\bar{p}_{c,i}$, i.e., $\bar{u}_{c,i}$ and $\bar{v}_{c,i}$, is calculated with~\eqref{equ:bar_pc_mc}. Furthermore, we leverage the following scheme to calculate $z_{c,i}$ (see Figure~\ref{fig_computation}):
\begin{figure}[!t]
	\centering
	\includegraphics[width=0.424\textwidth]{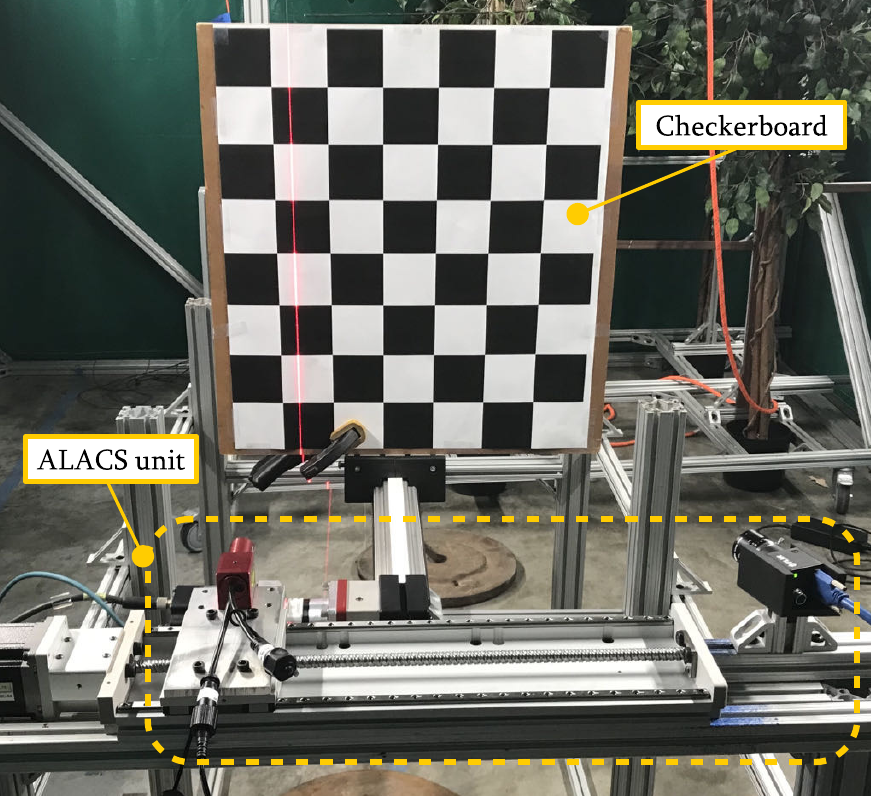}
	\caption{Hardware setup for extrinsic parameter calibration.}
	\label{fig_setup}
\end{figure}
	
\begin{enumerate}
	\item \textbf{Corner Detection.} The checkerboard corners are detected from the image by using the algorithm developed in~\cite{Geiger2012ICRA}. 
		
	\item \textbf{Pose Reconstruction.} Based on the detected checkerboard corners and the prior knowledge about the checkerboard square size, the relative pose information between the planar checkerboard and the camera is reconstructed~\citep{Hartley2003}. The pose information is described by the rotation matrix $R_{b}\in \mathbb{SO}^{3}$ and the translation vector $t_{b}\in \mathbb{R}^{3}$.
		
	\item \textbf{Computation of $z_{c,i}$}. Based on the relative pose information $R_{b}$, $t_{b}$ and the normalized coordinate $\bar{p}_{c,i}$, $z_{c,i}$ is calculated with projection geometry~\citep{Hartley2003}.
\end{enumerate}
To obtain multiple data samples $s_{i}=\begin{bmatrix}
	\bar{u}_{c,i}, \bar{v}_{c,i}, z_{c,i}
\end{bmatrix}^{\top}$ $(i=1, 2, \cdots, n)$, the planar checkerboard is moved to different positions, and an image is recorded at each position. For each image, several laser points are selected and the corresponding data samples $s_{i}=\begin{bmatrix}
	\bar{u}_{c,i}, \bar{v}_{c,i}, z_{c,i}
\end{bmatrix}^{\top}$ are computed by using the aforementioned strategy. A total of $n$ data samples will be collected and then used for the calibration of extrinsic parameters. 
	
\begin{algorithm}[!t]
	\caption{RANSAC-based robust calibration}\label{alg:calibration}
	\begin{algorithmic}
		\Require $\mathcal{S}=\left\lbrace s_{1}, s_{2}, \cdots, s_{n} \right\rbrace$, $k_{max}$, $\epsilon$
		\Ensure $\hat{\alpha}$, $\hat{L}_{0}$, $\hat{\beta}$
		\State $k=0$, $I_{max}=0$
		\While{$k < k_{max}$}
		\State \textbf{1. Hypothesis generation}
		\State Randomly select 4 data samples from $\mathcal{S}$ to construct the subset $\mathcal{S}_{k}=\left\lbrace s_{k_1}, s_{k_2}, s_{k_3}, s_{k_4} \right\rbrace$, where $\left\lbrace k_1, k_2, k_3, k_4 \right\rbrace \subset \left\lbrace 1, 2, \cdots, n \right\rbrace$
		\State Estimate parameters $\left(\hat{\alpha}_{k}, \hat{L}_{0,k}, \hat{\beta}_{k}\right)$ based on $\mathcal{S}_{k}$ and \eqref{equ:calibration_opt1}
		\State \textbf{2. Verification}
		\State Initialize the inlier set $\mathcal{I}_{k}=\left\lbrace \right\rbrace$
		\For{$i=1, 2, \cdots, n$}
		\If{$\left| z_{c,i} - \frac{\hat{L}_{0,k}}{\sin(\hat{\alpha}_{k}) - \bar{u}_{c,i}\cos(\hat{\alpha}_{k}) - \bar{v}_{c,i}\tan(\hat{\beta}_{k})}\right| \le \epsilon$}
		\State Add $s_{i}$ to the inlier set $\mathcal{I}_{k}$
		\EndIf
		\EndFor
		\If{$\left| \mathcal{I}_{k} \right| > I_{max}$ }
		\State $\mathcal{I}^{*} = \mathcal{I}_{k}$, $I_{max} = \left| \mathcal{I}_{k} \right|$
		\EndIf
		\State $k=k+1$
		\EndWhile
		\State Estimate parameters $\left(\hat{\alpha}, \hat{L}_{0}, \hat{\beta}\right)$ based on $\mathcal{I}^{*}$ and \eqref{equ:calibration_opt1}
	\end{algorithmic}
\end{algorithm}
	
In the second step, the extrinsic parameters are to be identified based on the model~\eqref{equ:zc} and the collected data samples $s_{i}=\begin{bmatrix}
	\bar{u}_{c,i}, \bar{v}_{c,i}, z_{c,i}
\end{bmatrix}^{\top}$ $(i=1, 2, \cdots, n)$. In the ideal case, each data sample $s_{i}$ should satisfy the relation~\eqref{equ:zc}. According to this observation, the extrinsic parameters $\alpha$, $L_{0}$, and $\beta$ can be estimated by solving the following optimization problem:
\begin{equation} \label{equ:calibration_opt1}
	\begin{aligned}
		& \min_{\hat{\alpha}, \hat{L}_{0}, \hat{\beta}} f = \sum_{i=1}^{n}\left( z_{c,i} -  \hat{z}_{c,i} \right)^{2},
		\\
		& \text{s.t.} \quad \hat{z}_{c,i} = \frac{\hat{L}_{0}}{\sin(\hat{\alpha}) - \bar{u}_{c,i}\cos(\hat{\alpha}) - \bar{v}_{c,i}\tan(\hat{\beta})}, 
		\\
		& \qquad \; i=1, 2, \cdots, n,
	\end{aligned}
\end{equation}
where $\hat{\alpha}$, $\hat{L}_{0}$, and $\hat{\beta} \in \mathbb{R}$ are estimated values of $\alpha$, $L_{0}$, and $\beta$, respectively.
Note that the minimization problem~\eqref{equ:calibration_opt1} directly applies all data samples to compute extrinsic parameters, which is not robust in the presence of data outliers. In general, the data samples $s_{i}=\begin{bmatrix}
	\bar{u}_{c,i}, \bar{v}_{c,i}, z_{c,i}
\end{bmatrix}^{\top}$ $(i=1, 2, \cdots, n)$ are corrupted with noises and may contain outliers that do not satisfy the relation~\eqref{equ:zc}. These outliers can severely influence the calibration accuracy and thus need to be removed. Towards that end, we adopt the random sample consensus (RANSAC) methodology \citep{FischlerCACM1981,RaguramTPAMI2013} to extract credible data from $\mathcal{S} = \left\lbrace s_{1}, s_{2}, \cdots, s_{n} \right\rbrace$.
The RANSAC-based robust calibration scheme is detailed in Algorithm~\ref{alg:calibration}. Specifically, the calibration scheme is divided into three steps. First, subsets of $\mathcal{S}$ are randomly selected to calculate different possible solutions to problems \eqref{equ:calibration_opt1}. Each one of these possible solutions is called a hypothesis in the RANSAC algorithm. Second, hypotheses are scored using the data points in $\mathcal{S}$, and the hypothesis that obtains the best score is returned as the solution. Finally, the data points that voted for the solution are categorized as a set of inliers and will be used to calculate the final solution.
	
The developed calibration scheme leverages RANSAC techniques to iteratively estimate the model parameters and select the solution with the largest number of inliers. Therefore, it is able to robustly identify the model parameters when some data samples are corrupted or noisy. 
	
\section{Experiments} \label{sec:results}
	
	
\subsection{Calibration Methods and Results}
As shown in Figure~\ref{fig_setup}, the experimental setup mainly consists of a specially designed ALACS unit and a planar checkerboard. To collect data samples for calibration, the planar checkerboard is placed in sequence at 10 different positions between 0.6 and 1.2 m from the ALACS unit, and at each position the FLIR camera is triggered to capture an image. For each image, 3 laser points are selected and the corresponding data samples $s_{i}=\begin{bmatrix}
	\bar{u}_{c,i}, \bar{v}_{c,i}, z_{c,i}
\end{bmatrix}^{\top}$ are computed by using the strategy introduced in Section~\ref{subsec:calibrationScheme}. A total of $n=30$ data samples are collected and then used for the calibration of extrinsic parameters.
	
To better evaluate the effectiveness of the developed high-fidelity model and robust calibration scheme, four different methods are implemented and tested on the same data samples. These four methods are introduced, as follows:
	
\begin{itemize}
	\item \textbf{Method 1:} This method utilizes the low-fidelity model to conduct the calibration. Specifically, the low-fidelity model only considers two extrinsic parameters $\alpha$ and $L$ and assumes that $\beta=0$. Under this case, the depth measurement mechanism of the ALACS unit degenerates into
	\begin{equation} \label{equ:zc_low}
		\begin{aligned}
			z_{c,i} = \frac{L}{\sin(\alpha) - \bar{u}_{c,i}\cos(\alpha)}. 
		\end{aligned}
	\end{equation} 
	The model \eqref{equ:zc_low} and all collected data samples are used to estimate the extrinsic parameters $\alpha$ and $L$.
		
	\item \textbf{Method 2:} Both the low-fidelity model~\eqref{equ:zc_low} and RANSAC techniques are used for calibration. Compared with Method 1, this method leverages RANSAC to remove outlier data.
		
	\item \textbf{Method 3:} This method computes the extrinsic parameters $\alpha$, $L_{0}$, and $\beta$ by solving the optimization problem~\eqref{equ:calibration_opt1}, which is designed based on the high-fidelity model~\eqref{equ:zc} and all data samples.  
		
	\item \textbf{Method 4:} This is our developed method which combines the high-fidelity model with RANSAC techniques for calibration. The method is detailed in Algorithm~\ref{alg:calibration}.
\end{itemize}
	
\begin{table*}[!h]
	\centering
	\begin{tabular}{lcccc}
		\hline
		& $\alpha$ (deg) & $L_{0}$ (mm) & $\beta$ (deg) & Mean Error $\left| z_{c,i}-\hat{z}_{c,i} \right|$ (mm)
		\\ \hline
		Method 1 (Low-fidelity model + All data)& 19.03 & 382.83 & / & 4.91 
		\\
		Method 2 (Low-fidelity model + RANSAC) & 19.28 & 386.37 & / & 3.80 
		\\
		Method 3 (High-fidelity model + All data) & 19.01 & 381.09 & 0.73 & 1.84
		\\
		\textbf{Method 4 (High-fidelity model + RANSAC)} & 19.07 & 381.98 & 0.69 & \textbf{0.39}
		\\ \hline       
	\end{tabular}
	\caption{Calibration results by using four methods.}
	\label{table:calibration}
\end{table*}
	
The mean error of $\left| z_{c,i}-\hat{z}_{c,i} \right|$ is computed to evaluate the performance of these four methods. The calibration results are summarized in Table~\ref{table:calibration}. Both Methods 1 and 2 use model~\eqref{equ:zc_low} for calibration, while Methods 3 and 4 rely on model~\eqref{equ:zc}. From Table~\ref{table:calibration}, it can be seen that Methods 3 and 4 achieve better calibration performance than Methods 1 and 2, indicating that the high-fidelity model~\eqref{equ:zc} can well pair the laser with the RGB camera for depth measurements. Moreover, by comparing Method 3 with Method 4, it can be concluded that the RANSAC technique is robust for removal of outlier data and the developed calibration method is effective in determining the extrinsic parameters of the ALACS unit.
	
\subsection{Localization Accuracy}
As mentioned in Section~\ref{subsec:calibrationScheme}, the parameters $\alpha$ and $\beta$ are constants, while $L$ is variable since the laser position can be adjusted via the linear motion slide. The linear motion slide is fixed at an initial position (i.e., $L$ is fixed to $L_{0}$) during the calibration procedure.
We change the value of $L$ by moving the laser to different positions and collect data samples to fully evaluate the localization accuracy of the ALACS unit. More precisely, the laser is moved from its initial position towards the camera side by $d$ cm, where $d$ is selected as the following values in turn: 
\[
d = 0, 5, 10, 15, 20.
\]
Given $L_{0}$ and $d$, $L$ can be computed by $L=L_{0}-d$. For each laser position (i.e., for each $L$ value), 10 images are collected with the planar checkerboard being placed at different positions between 0.6 and 1.2 m away from the ALACS unit. 3 laser points are randomly chosen from each image, and then at each laser position, a total of 30 data samples are utilized to evaluate the localization accuracy of the ALACS unit. The 3D measurements of the collected data, i.e., $p_{c,j}=\begin{bmatrix}
	x_{c,j}, y_{c,j}, z_{c,j}
\end{bmatrix}^{\top}$ ($j=1, 2, \cdots, 30$), are obtained with the aid of the checkerboard setup. Meanwhile, the extrinsic parameters calculated with the developed robust calibration scheme (see Table~\ref{table:calibration}) are used to determine the estimated 3D measurements $\hat{p}_{c,j}=\begin{bmatrix}
	\hat{x}_{c,j}, \hat{y}_{c,j}, \hat{z}_{c,j}
\end{bmatrix}^{\top}$.  

\begin{figure*}[!h]
	\centering
	\begin{subfigure}[b]{1\textwidth}
		\centering
		\includegraphics[width=1\textwidth]{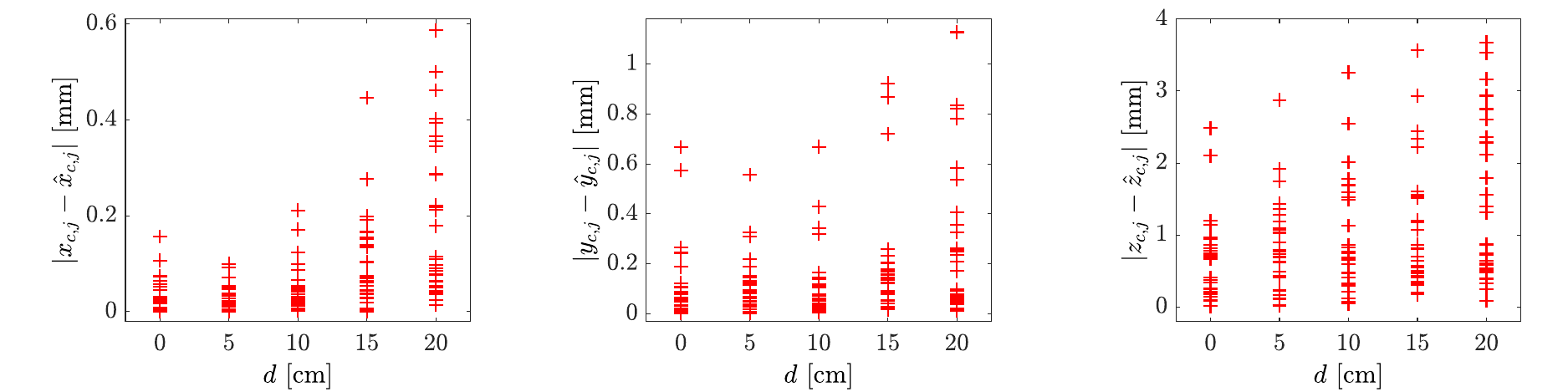}
		\caption{}
		\label{fig_exp1}
	\end{subfigure}
	\hfill
	\begin{subfigure}[b]{1\textwidth}
		\centering
		\includegraphics[width=1\textwidth]{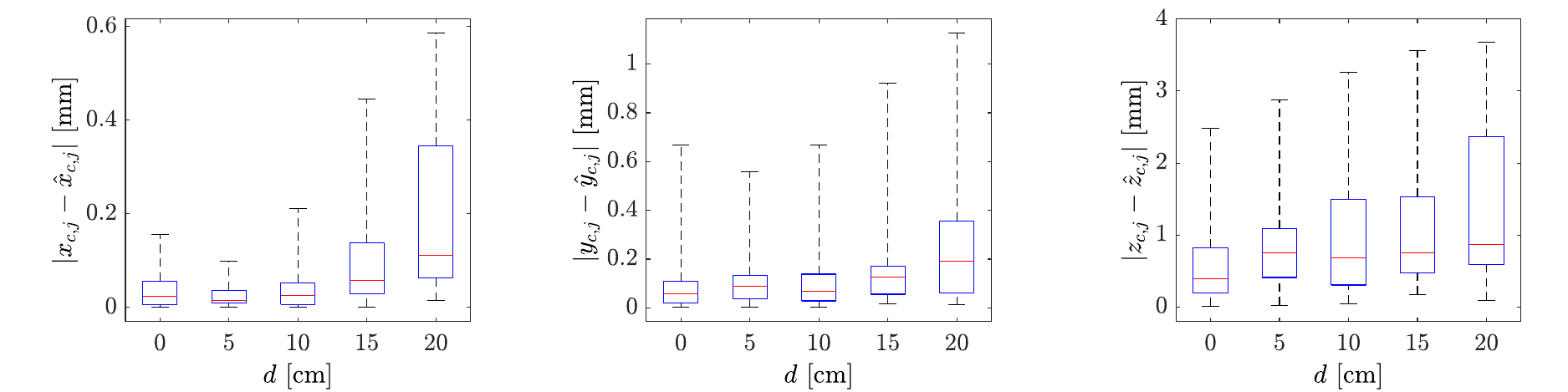}
		\caption{}
		\label{fig_exp2}
	\end{subfigure}
	\caption{Localization accuracy of ALACS when the laser is adjusted to different positions (i.e., $d = 0, 5, 10, 15, 20$ cm). (a) Localization error distribution at 5 different laser positions. (b)   Statistics summary of the localization error distribution. On each box, the central red mark is the median, the edges of the box are the 25th and 75th percentiles, the whiskers extend to the most extreme data points.}
	\label{fig_exp}
\end{figure*}
	
The localization results are shown in Figure~\ref{fig_exp}. Specifically, Figure~\ref{fig_exp1} shows the localization error distribution of ALACS with laser being placed at 5 different positions, and Figure~\ref{fig_exp2} depicts the corresponding statistical metrics. It can be found from the results that the ALACS unit achieves precise localization in the $x$ (horizontal), $y$ (vertical), and $z$ (depth) directions. In most instances, the localization errors along $x$, $y$, and $z$ directions are within $0.4$ mm, $0.8$ mm, and $3$ mm, respectively. Even under the worst-case scenarios, the largest localization errors along these three directions are less than $0.6$ mm, $1.2$ mm, and $4$ mm, respectively, when the distance between the planar checkerboard and ALACS is within 0.6$\sim$1.2 m. Note that our robotic harvesting system uses a vacuum-based end-effector to grasp and detach fruits, and the end-effector is able to attract fruits within a distance of about $1.5$ cm. Therefore, according to the evaluation results, it can be concluded that the ALACS unit can meet the requirements for fruit localization and can be integrated with other hardware modules for automated apple harvesting. 

The RealSense D435i RGB-D camera was used in our previous apple harvesting robotic prototypes to localize the fruit~\citep{Zhang2021MECH,Zhang2022IROS}. 
According to the manufacturer's datasheet~\citep{RealSense}, this camera offers a measurement accuracy of less than 2\% of the depth range. This suggests that the maximum localization error along the depth direction is estimated to be less than 24 mm within the distance range of 0.6 to 1.2 m between the target and the camera. On the other hand, the ALACS unit demonstrates a maximum depth measurement error of 4 mm at distance ranging from 0.6 to 1.2 m. These results indicate that the ALACS unit has promising potential for achieving precise and reliable fruit localization.
	
\section{Conclusion} \label{sec:conclu}
This paper has reported the system design and calibration scheme of a new perception module, called Active LAser-Camera Scanner (ALACS), for fruit localization. A red line laser, an RGB camera, and a linear motion slide were fully integrated as the main components of the ALACS unit. A high-fidelity model was established to reveal the localization mechanism of the ALACS unit. Then, a robust scheme was proposed to calibrate the model parameters in the presence of data outliers. Experimental results demonstrated that the proposed calibration scheme can achieve accurate and robust parameter computation, and the ALACS unit can be exploited for localization with the maximum errors being less than $0.6$ mm, $1.2$ mm, and $4$ mm in the horizontal, vertical, and depth directions, respectively, when the distance between the target and ALACS is within 0.6$\sim$1.2 m. Future work will include further improvements on the efficiency of the scanner such that it can provide a faster measurement to support multiple arms planned in our next version of the harvesting robot. In addition, we will design comprehensive experiments to compare the measurement accuracy of the ALACS unit and consumer depth cameras.
	
\section*{Authorship Contribution}
\textbf{Kaixiang Zhang}: Formal analysis, Software, Writing - original draft;
\textbf{Pengyu Chu}: Software, Writing - review  \& editing;
\textbf{Kyle Lammers}: Software, Writing - review  \& editing; 
\textbf{Zhaojian Li}: Supervision, Resources, Writing - review \& editing;   
\textbf{Renfu Lu}: Supervision, Resources, Writing - review \& editing.
	
\section*{Acknowledgement}
This research was supported by the U.S. Department of Agriculture Agricultural Research Service. The findings and conclusions in this paper are those of the authors and should not be construed to represent any official USDA or U.S. Government determination or policy. Mention of commercial products in the paper does not imply endorsement by USDA over those not mentioned.
	
\typeout{}
\bibliography{ref}
	
\end{document}